# Modeling Spatial and Temporal Cues for Multi-label Facial Action Unit Detection

Wen-Sheng Chu, Fernando De la Torre, and Jeffrey F. Cohn

Robotics Institute, Carnegie Mellon University

**Abstract.** Facial action units (AUs) are essential to decode human facial expressions. Researchers have focused on training AU detectors with a variety of features and classifiers. However, several issues remain. These are *spatial representation*, *temporal modeling*, and *AU correlation*. Unlike most studies that tackle these issues separately, we propose a hybrid network architecture to jointly address them. Specifically, spatial representations are extracted by a Convolutional Neural Network (CNN), which, as analyzed in this paper, is able to reduce person-specific biases caused by hand-crafted features (*e.g.*, SIFT and Gabor). To model temporal dependencies, Long Short-Term Memory (LSTMs) are stacked on top of these representations, regardless of the lengths of input videos. The outputs of CNNs and LSTMs are further aggregated into a fusion network to produce per-frame predictions of 12 AUs. Our network naturally addresses the three issues, and leads to superior performance compared to existing methods that consider these issues independently. Extensive experiments were conducted on two large spontaneous datasets, GFT and BP4D, containing more than 400,000 frames coded with 12 AUs. On both datasets, we report significant improvement over a standard multi-label CNN and feature-based state-of-the-art. Finally, we provide visualization of the learned AU models, which, to our best knowledge, reveal how machines see facial AUs for the first time.

## 1 Introduction

Facial actions convey information about a person's emotion, intention, and physical state, and are vital for use in studying human cognition and related processes. To encode such facial actions, the Facial Action Coding System (FACS) [10] is the most comprehensive. With FACS, progress was enabled in affective computing, social signal processing and behavioral science. FACS segments visual effects of facial activities into action units (AUs), which has shown a powerful description in universal expressions and led discoveries to many areas such as marketing, mental health, and entertainment.

In computer vision, a conventional pipeline of automated facial AU detection compiles four major stages: detection ↦ alignment ↦ representation ↦ classification. With the progress made in face detection and alignment, most research nowadays focuses on features, classifiers, or their combinations. However, due to slow-growing rate in the amount of FACS-coded data, it remains unclear how to pick the best combination that generalizes across subjects and datasets. At least three issues arise in the literature of automated AU detection: (1) *Spatial representation:* Engineered features, *e.g.*, SIFT, have shown to cause person-specific biases in estimating AUs, causing sophisticated learning methods such as personalization [3, 27, 38]. A good representation for AUs must generalize to unseen subjects, regardless of the existence of individual differences caused by appearance, behaviors or facial morphology. (2) *Temporal modeling:* Temporal information is crucial for telling AUs like humans. Due to the richness, ambiguity, and dynamic nature of facial actions, it remains unclear how such temporary memory can be effectively encoded and recalled. (3) *AU correlation:* The presence of AUs influences each other. Fig. 1(a) illustrates an example of likely and unlikely co-occurring AUs. For instance, the occurrence of AU12 suggests an occurrence of AU6, and reduces the likelihood of AU15. Such correlation helps an AU detector determine one AU given others. Despite the seemingly unrelated nature of the three issues, it is possible to consider them complementarily. That is, a good representation would assist learning temporal models and AU correlations, and knowing AU correlations could benefit representation learning and temporal modeling. Most existing studies, however, explore these issues separately, and, therefore, are not able to fully capture their entangled dependencies.

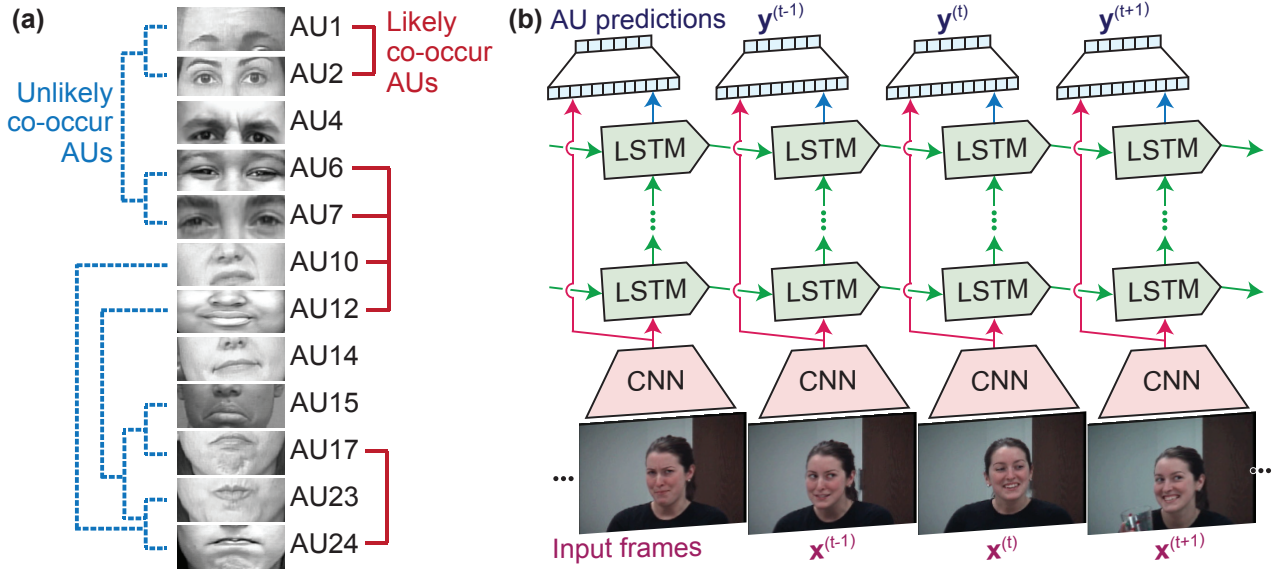

**Fig. 1.** An overview of the proposed hybrid deep learning framework: (a) An illustration of AU relations, showing the multi-label nature of AU detection. (b) The proposed network possesses both strengths of CNNs and LSTMs to model and utilize both spatial and temporal cues.

To address the above issues, this paper proposes a hybrid network architecture that models both spatial and temporal relationships from jointly multiple AUs. The proposed network is appealing for naturally modeling the three complementary aspects of AU data. Fig. 1 gives an overview of the proposed framework. To learn a generalizable representation, a CNN is trained to extract spatial features. As analyzed in this study, such features reduce person-specific biases that were identified in hand-crafted features [3, 27, 38], and thus offer possibilities to reduce the burden of designing sophisticated classifiers. To capture temporal dependencies, LSTMs are stacked on top of the spatial features. We aggregate the output scores from both CNNs and LSTMs into a fusion network to produce a per-frame prediction for 12 AUs. Extensive experiments were performed on two spontaneous AU datasets, GFT and BP4D, containing totally >400,000 frames. We report that the learned spatial features, further combined with temporal information, outperform a standard CNN and feature-based state-of-the-art methods. In addition, we visualize notions of different AUs captured by the model, which, to our best knowledge, reveal how machines see face AUs for the first time.

## 2   Related Work

This study is motivated by contemporary issues in automated facial AU detection and success in deep networks. Below we review both fields and build their connections.

**Facial AU detection:** With significant progress made in face detection and alignment, most prior work focus on developing features, classifiers, and their fusions. See [6,25,28,34] for surveys. Three are recent issues risen in automated AU detection.

First is *spatial representation*, which is typically biased to individual differences in appearance, behavior, recording environments, etc. This factor produces shifted distributions in feature space, and thus hinders the generalizability of pre-trained classifiers. To reduce such distribution mismatch, a few studies merged to *personalization* techniques. Chu *et al.* [3] proposed to personalize a generic classifier by iteratively reweighting training samples based on relevance to a test subject. Along this line, Sangineto *et al.* [27] directly transferred classifier parameters from a set of source subjects to a test one. Zeng *et al.* [41] adopted an easy-to-hard strategy by propagating predictions from confident classifiers that are trained on samples with high prediction scores. Yang *et al.* [38] further extended personalization for estimating AU intensities by removing a person's identify with a latent factor model. Rudovic *et al.* [26] interpreted the person-specific variability as a context-modeling problem, and propose a conditional ordinal random field to address context effects. Others sought to learn AU-specific facial patches to specialize the representation [23,43,44]. However, while progress has been made, these studies still resort to hand-crafted features. We argue that person-specific biases from such features can be instead reduced by learning them.

Another issue remains in *temporal modeling*, as modeling dynamics is crucial in recognizing actions like humans. To explore the temporal context, graphical models have been popularly used for AU detection. Non-parametric HMMs [29] were introduced to encode discrimination ability at class and state levels. A hidden CRF [1] classified over a sequence and established connections between the hidden states and AUs. However, these models made Markov assumption and thus lacked consideration of long-term dependencies. Switching Gaussian process models [2] was built upon dynamic systems and Gaussian process to simultaneously track motions and recognize events. The Gaussian assumption unnecessarily holds in real-world scenarios where we do not know from which distribution video frames are sampled. In this paper, we attempt to learn long-term dependencies to improve predicting AUs without the requirement to a priori of state dependencies and distributions.

Last but not least, AUs retrain *correlations*, which make itself a problem different from standard expression recognition, *e.g.*, [9,21,22]. To capture such correlation, a generative dynamic Bayesian networks (DBN) [33] was proposed to model the AU relationships and their temporal evolutions. Rather than learning, pairwise AU relations can be explicitly inferred using statistics in annotations, and then injected such relations into a multi-task learning framework to select important patches for each AU [43]. In addition, a restricted Boltzmann machine (RBM) [35] was developed to directly capture the dependencies between image features and AU relationships. Following this direction, image features and AU outputs were fused in a continuous latent space using a conditional latent variable model [11]. Although improvements can be observed from jointly predicting multiple rather than individual AUs, these approaches rely on engineered features such as SIFT, LBP, or Gabor. Table 1 summarizes the comparisons.

**Table 1.** Comparisons between this study and alternative AU detection methods

| AU detection methods | Spatial representation | Temporal modeling | AU correlation |
|---|---|---|---|
| [3, 23, 26, 27, 41, 43, 44] | ✓ | ✗ | ✗ |
| [1, 2, 7, 15, 17, 29] | ✗ | ✓ | ✗ |
| [11, 12, 21, 22, 33, 35, 43] | ✗ | ✗ | ✓ |
| The proposed method | ✓ | ✓ | ✓ |

**Deep networks:** Recent success of deep networks suggests that strategically composing layers of nonlinear functions can result in powerful models for perceptual problems. Closest to our work are the ones in facial AU detection and video classification.

Most deep networks for AU detection directly adapt CNNs. Gadi *et al.* [14] used a 7-layer CNN for estimating AU occurrence and intensity. AU-aware deep networks [21] learned representation directly from images, and then greedily picked relevant receptive fields according to a relevance measure. Ghosh *et al.* [12] showed that a shared representation can be directly learned from input images using a multi-label CNN. However, no temporal context was involved in learning these networks. To incorporate temporal modeling, Jaiswal *et al.* [15] trained CNNs and BLSTM on shape and landmark features to predict for individual AUs. Because input features were predefined masks and image regions, unlike this study, gradient cannot backprop to full face image to analyze the per-pixel contribution to each AU. In addition, it ignored dependencies between AUs and a multi-modal fusion that could improve performance in video prediction, *e.g.*, [31, 37]. On the contrary, our network simultaneously models spatial-temporal context and AU dependencies, and thus serves as a more natural framework for AU detection.

The construction of our network is inspired by studies in video classification. Simonyan *et al.* [31] proposed a two-stream CNN that captures information from static frames and motion optical flow between frames. A video class was predicted by fusing scores from both networks using either average pooling or an additional SVM. To incorporate "temporally deep" models, Donahue *et al.* [8] proposed a general recurrent convolutional network that combines both CNNs and LSTMs, which can be then specialized into tasks such as activity recognition, image description and video description. Similarly, Wu *et al.* [37] used both static frames and motion optical flow, combined with two CNNs and LSTMs, to perform video classification. Video-level features and LSTM outputs were fused to produce a per-video prediction. Our approach fundamentally differs from the above networks in several aspects: (1) Video classification is a multi-class problem, yet AU detection is multi-label. (2) Motion optical flow is useful in video classification, but not in AU detection due to large head movements. (3) AU detection requires per-frame detection; video classification is video-based prediction.

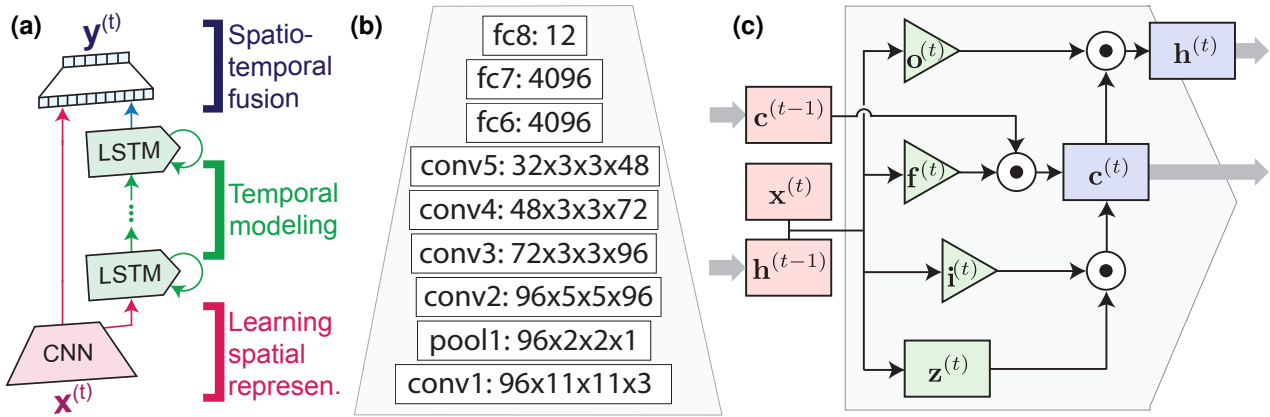

**Fig. 2.** The structure of the proposed hybrid network: (a) Folded illustration of Fig. 1, showing 3 components of learning spatially representation, temporal modeling, and spatiotemporal fusion, (b) our 8-layer CNN architecture, and (c) the schematic of an LSTM block.

## 3 The Hybrid Network for Multi-label AU Detection

In this section, we describe the hybrid network for multi-label AU detection. Fig. 2(a) shows a folded illustration of the network, which composes three building components: learning spatial representation with CNNs, temporal modeling with LSTMs, and frame-based spatiotemporal fusion. Below we describe each component in turn.

### 3.1 Learning spatial representation

In AU detection, one challenge remains that face shape and appearance undermine generalization of AU detectors across subjects [3, 27, 38]. We argue that a specialized representation could be learned to reduce the burden of designing a sophisticated classifier, and further improve detection performance. In addition, AUs are correlated: Some AUs are known to co-occur frequently (*e.g.*, AUs 6+12 in a Duchenne smile), and some infrequently. Knowing such AU relations, as illustrated in Fig. 1(a), is likely to lead to a more reliable classifier [11, 43]. To this end, we train a multi-label convolutional neural network (CNN) due to its proven power in learning features and superior performance in classification. This network aims to model dependencies between AUs, and meanwhile discriminates one AU from another. Here we modified AlexNet [19] and designed an 8-layer architecture, as shown in Fig. 2(b). Given a ground truth label $\mathbf{y} \in \{-1, 0, 1\}^L$ and a prediction $\widehat{\mathbf{y}} \in \mathbb{R}^L$ for $L$ AU labels, this multi-label CNN aims to minimize the multi-label cross entropy loss:

$$L_E(\mathbf{y}, \widehat{\mathbf{y}}) = \frac{-1}{L} \sum_{\ell=1}^{L} [y_\ell > 0] \log \widehat{y}_\ell + [y_\ell < 0] \log(1 - \widehat{y}_\ell), \tag{1}$$

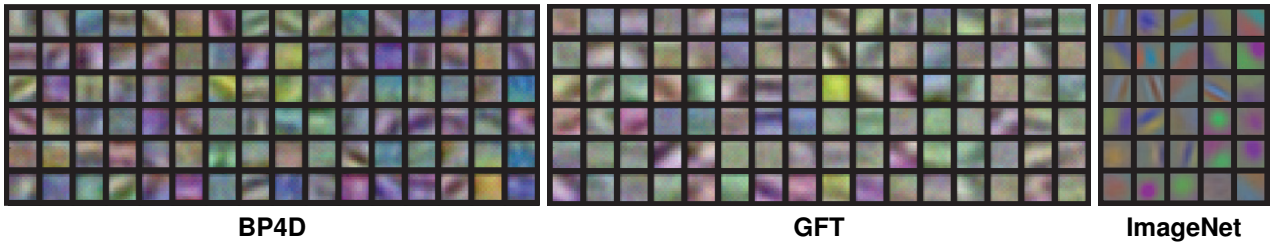

**Fig. 3.** `conv1` kernel visualization on BP4D, GFT, and ImageNet [19] (selected color blob detectors). As can be seen, filters learned on face datasets (BP4D and GFT) contain less color blob detectors, suggesting color information is less useful in the AU detection task. (best view in color)

where $[x]$ is an indicator function returning 1 if $x$ is true, and 0 otherwise. The outcome from the fc7 layer is extracted and $L_2$ normalized as the final representation, resulting in a 4096-D vector. We denote this representation "fc7" hereafter. Due to dropout regularization and ReLu, fc7 feature contains ∼35% zeros out of 4096 values, making it significantly sparser compared to standard engineered features such as SIFT or Gabor. The proposed multi-label CNN is similar to [12] and AlexNet [19], with slightly different architecture and purpose. [12] takes a 40×40 image as input, which to our experience, could be insufficient for recognizing the subtle AUs on faces. AlexNet was originally designed for object classification, yet for face images, the color information could be less useful than for natural images such as objects and scenes. Instead, we train the entire network from scratch. Fig. 3 visualizes the learned kernels from the `conv1` layer on both AU datasets (BP4D and GFT) and ImageNet [19]. As can be seen, the kernels learned on GFT and BP4D contain less color blob detectors than the ones learned on ImageNet. In Sec. 4, we will empirically show that fc7 is more robust against individual differences compared to hand-crafted features such as SIFT or Gabor.

### 3.2 Temporal modeling with stacked LSTMs

It is usually hard to tell an "action" by looking at only a single frame. Fig. 4 illustrates a common scenario in facial AU detection. Without observing previous frames, one could find it difficult to tell whether a person feels depressed or just being under transition of an expression. Additionally, as can be seen, the existence of common temporal pattern suggests strong temporal dependencies. This section aims to model such dependencies.

Having fc7 extracted, we use a stacked LSTM architecture [13] for temporal modeling, as illustrated in Fig. 1. Fig. 2(c) shows the schematic of an LSTM block. We experimented various numbers of layers and memory cells, and chose 3 stacks of LSTMs with 256 memory cells each. Because AUs involve much ambiguity and dynamics, the transition between two frames could encode crucial information for prediction. Unlike learning spatial representation on fixed and cropped images, videos can span widely in temporal context and be difficult to be modeled with a fixed-size architecture, *e.g.*, [1, 18, 29]. LSTM serves as an ideal model for avoiding the well-known "vanishing gradient" effect in recurrent models, and makes it possible to model long-term dependencies.

**Recurrent LSTMs:** Denote a sequence of input frames as $(\mathbf{x}^{(1)}, \ldots, \mathbf{x}^{(T)})$, and their labels as $(\mathbf{y}^{(1)}, \ldots, \mathbf{y}^{(T)})$, where superscripts indicate time steps. A recurrent model is expressed by iterating the equations from $t = 1$ to $T$:

$$\mathbf{h}^{(t)} = \mathcal{H}(\mathbf{W}_{xh}\mathbf{x}^{(t)} + \mathbf{W}_{hh}\mathbf{h}^{(t-1)} + \mathbf{b}_h), \tag{2}$$

$$\mathbf{y}^{(t)} = \text{softmax}(\mathbf{W}_{hy}\mathbf{h}^{(t)} + \mathbf{b}_y), \tag{3}$$

where $\mathbf{W}$ denotes weight matrices, $\mathbf{b}$ denotes bias vectors, $\mathcal{H}$ is the hidden layer activation function (typically the logistic sigmoid function), and the subscripts $\{x, h, y\}$ denote the (input,hidden,output) layers respectively. LSTM replaces the hidden nodes in the recurrent model with a memory cell, which allows the recurrent network to remember long term context dependencies. Given an input vector $\mathbf{x}^{(t)}$ at each time $t$ and the hidden state from previous time $\mathbf{h}^{(t-1)}$, we denote a linear mapping as:

$$\phi_\star^{(t)} = \mathbf{W}_\star \mathbf{x}^{(t)} + \mathbf{R}_\star \mathbf{h}^{(t-1)} + \mathbf{b}_\star, \tag{4}$$

where $\mathbf{W}$ is the rectangular input weight matrices, $\mathbf{R}$ is the square recurrent weight matrices, and $\star$ denotes one of LSTM components $\{c, f, i, o\}$, *i.e.*, cell unit, forget gate, input gate, and output gate. Element-wise activation functions are applied to introduce nonlinearity. Gate units often use a *logistic sigmoid* activation $\sigma(a) = \frac{1}{1+e^{-a}}$; cell units are transformed with *hyperbolic tangent* $\tanh(\cdot)$. Denote the point-wise multiplication of two vectors as $\odot$, LSTM applies the following update operations: (block input) $\mathbf{z}^{(t)} = \tanh(\phi_c^{(t)})$, (forget gate) $\mathbf{f}^{(t)} = \sigma(\phi_f^{(t)})$, (input gate) $\mathbf{i}^{(t)} = \sigma(\phi_i^{(t)})$, (output gate) $\mathbf{o}^{(t)} = \sigma(\phi_o^{(t)})$, (cell state) $\mathbf{c}^{(t)} = \mathbf{i}^{(t)} \odot \mathbf{z}^{(t)} + \mathbf{f}^{(t)} \odot \mathbf{c}^{(t-1)}$, and (block output) $\mathbf{h}^{(t)} = \mathbf{o}^{(t)} \odot \tanh(\mathbf{c}^{(t)})$. As seen in the update of cell states, an LSTM cell involves *summation* over previous cell states. The gradients are distributed over sums, and propagated over a longer time before vanishing. Because AU detection is by nature a *multi-label classification* problem, we optimize LSTMs to jointly predict multiple AUs according to the maximal-margin loss:

$$L_M(\mathbf{y}, \widehat{\mathbf{y}}) = \frac{1}{n_0} \sum_i \max(0, \lambda - y_i \widehat{y}_i), \tag{5}$$

where $\lambda$ is a pre-defined margin, and $n_0$ indicates the number of non-zero elements in ground truth $\mathbf{y}$. Although a typical $\lambda$ is set to be 1 (such as in regular SVMs), here we empirically choose $\lambda = 0.5$ because the activation function has squeezed the outputs into $[-1, 1]$, making the prediction value never go beyond $\lambda = 1$. During back propagation, we pass to previous layers the gradient $\frac{\partial L}{\partial \widehat{y}_i} = -\frac{y_i}{n_0}$ if $y_i \widehat{y}_i < 1$, and $\frac{\partial L}{\partial \widehat{y}_i} = 0$ otherwise. At each time step, LSTMs output a vector indicating potential AUs.

**Practical issues:** There has been evidence that using deep LSTM structure preserves better descriptive power than a single-layer LSTM [13]. However, because fc7 features are 4096-D, our design of LSTMs can lead to a very large model of $>1.3$ million parameters. To ensure that the number of parameters and the size of our datasets maintains the same order of magnitude, we applied PCA to reduce the fc7 features to 1024-D (about 98% energy preserved). We set dropout rate as 0.5 to the input and hidden layers, resulting in a final model of $\sim0.2$ million parameters, which turns out to work reasonably on our datasets. More implementation details are in Sec. 4.

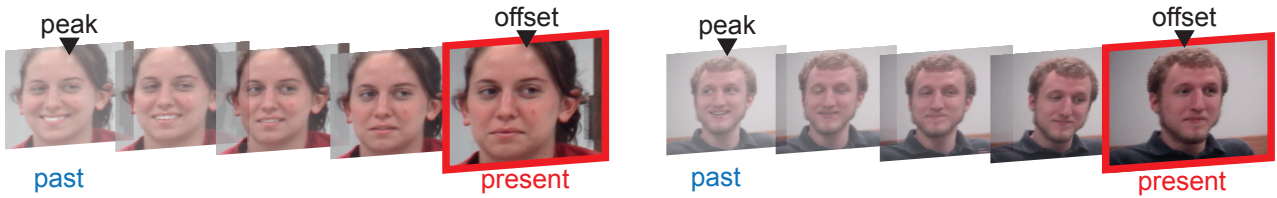

**Fig. 4.** A temporal pattern of subjects G128B1 (left) and G060A1 (right) from GFT: A transition from an AU12 peak frame to an offset frame, showing the need of modeling temporal dependency.

### 3.3   Frame-based spatiotemporal fusion

The spatial CNN performs AU detection from still video frames, while the temporal LSTM is trained to detect AUs from temporal transitions. Unlike video classification that produces video-based prediction, we model the correlations between spatial and temporal cues by adding an additional fusion network. We modify the late fusion model [18] to achieve this goal. Fig. 1(b) gives an illustration. For each frame, two separate fully connected layers with shared parameters are placed on top of both CNNs and LSTMs. The fusion network merges the stacked $L_2$-normalized scores in the first fully connected layer. In experiments, we see this fusion approach consistently improves the performance compared to CNN-only results.

## 4   Experiments

### 4.1   Datasets

We evaluated the proposed hybrid network on two of the largest spontaneous datasets BP4D [42] and GFT [4]. Each dataset was FACS-coded by experienced FACS coders. Only AU occurring more than 5% base rate were included for analysis. In total, we selected 12 AUs to perform the experiments, resulting in >400,000 valid frames. Unlike previous studies that suffer from scalability issues and require downsampling of training data, the network is in favor of large dataset so we made use of all available data. Note that the CK+ benchmark [24] is not applicable because the AU annotations are given on single video; we aim at per-frame prediction.

**BP4D** [42] is a spontaneous facial expression dataset in both 2D and 3D videos. The dataset includes 41 participants aging from 18 to 29 associating with 8 tasks, which are covered with an interview process and a series of activities to elicit eight emotions. Frame-level ground-truth for facial actions are obtained using the FACS. In our experiments, we used 328 2D videos from 41 participants, resulting in 146,847 available frames with AU coded. We selected positive samples as those with intensities equal or higher than A-level, and negative samples as the remaining.

**GFT** [4] contains 720 participants recorded when three previously unacquainted young adults sat around a circular table for 30 minutes of conversation with drinks. Moderate out-of-plane head motion and occlusion are presented in the videos and make the AU detection challenging. We used 50 participants with each containing one video of about 2 minutes (~5000 frames), resulting in 254,451 available frames with AU coded. Frames with intensities equal or greater than B-level are used as positive, otherwise, intensities less than B-level are negative.

### 4.2   Settings

**Pre-processing:** We pre-processed all videos by extracting facial landmarks using the IntraFace software [5]. Tracked faces were registered to a reference face using similarity transform, resulting in $200 \times 200$ face images, which were then randomly cropped into $176 \times 176$ and/or flipped for data augmentation. Each frame was labeled +1/-1 if an AU is present/absent, and 0 otherwise (*e.g.*, lost face tracks or occluded face).

**Dataset splits:** For both datasets, we adopted two protocols. First is a *3-fold protocol*: Each dataset was evenly partitioned into 3 folds with exclusive subjects. We iteratively trained a model using two folds and evaluated on the remaining one, until all subjects were tested. Validation was assigned to ∼20% of the training subjects. To maximize the limit of deep models, we adopted an additional train/validation/test splits as in the deep learning literature (*e.g.*, [19, 31, 37]). Specifically, we used a *10-fold protocol*, where 9 folds were for training/validation and one fold for test. Different from the 3-fold protocol, here only the one out of 10 folds was tested. In addition, to measure the transferability of fc7 features, we performed a *cross-dataset* protocol by training CNNs on one dataset and using it to extract spatial representations on another.

**Evaluation metrics:** To provide an evaluation in an objective manner, we reported performance using three metrics[1]. Denote $R$ and $P$ as recall and precision. Frame-based F1-score (F1-frame$=\frac{2RP}{R+P}$) is used for its popularity in AU detection. It serves one gold standard to compare with results reported in the literature. To compensate the skewed nature of AUs, F1-norm computes a skew-normalized F1-frame by multiplying false negatives and true negatives by the factor of skewness, which is computed as the ratio of positive samples over negative ones. Because AUs occur as temporal signals, we also evaluated an event-based F1 (F1-event$=\frac{2ER \cdot EP}{(ER+EP)}$) to measure detection performance at segment-level, where $ER$ and $EP$ are event-based recall and precision as defined in [7]. Each metric captures different properties about the results, and thus is able to tell the prediction power in term of spatial and temporal consistency. For each method, we reported all metrics on each AU and their averages.

**Network settings and training:** We trained the CNNs with mini-batches of 196 samples, a momentum of 0.9 and weight decay of 0.0005. All models were initialized with learning rate of 0.001, which was further reduced manually whenever the validation error stopped decreasing. The implementation was based on the Caffe toolbox [16] with modification to support multi-label cross-entropy loss. For training LSTMs, we set an initial learning rate of 1e-3 in conjunction of momentum of 0.9, weight decay 0.97, and RMSProp for stochastic gradient descent. All gradients were computed using backpropagation through time (BPTT) on 10 randomly sampled sequences in parallel, each drawn from the training set. All sequences were 1300 frames long, and the first 10 frames were disregarded during the backward pass, as they may have insufficient temporal context. In the end, our network went through about 10 passes over the full training set. The matrices $\mathbf{W}$ were initialized within $[-0.08, 0.08]$. As AU data is heavily skewed, *i.e.*, some AUs occur rarely than others and only a sparse subset of AU occur at a time, randomly sampled the sequences could cause LSTMs biased to negative predictions. As a result, we omitted the sampled sequences with less than 1.5 active AUs per frame. All experiments were performed on one NVidia Tesla K40c GPU.

---

[1] Evaluation code: <https://github.com/l2ior/metrics>

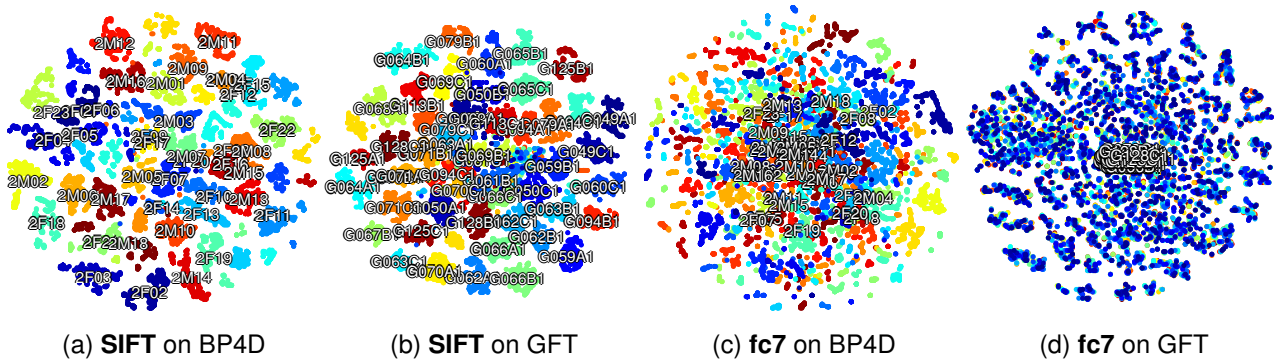

| (a) **SIFT** on BP4D | (b) **SIFT** on GFT | (c) **fc7** on BP4D | (d) **fc7** on GFT |

**Fig. 5.** A t-SNE embedding of SIFT and fc7 features on BP4D and GFT datasets. **(a)(b)** SIFT features colored in terms of *subjects*. **(c)(d)** fc7 features colored in AU 12. Each text represents one subject ID and is placed at the center of its own frames. The clustering effect in SIFT features reveal that face images retain individual differences; the learned fc7 reduces such influence.

### 4.3   Evaluation of learned representation

To answer the question whether individual differences can be reduced by feature learning, we first evaluated the fc7 features with standard features in AU detection, including both shape, Gabor, and SIFT features. Because such features for AU detection are unsupervised, for fairness, fc7 features for BP4D were extracted using CNNs trained on GFT, and vise versa. Fig. 5 shows the t-SNE embeddings of frames represented by SIFT and fc7 features, and visualize the effect of individual differences by coloring in terms of subjects. As can be seen in (a) and (b), SIFT has strong distributional biases where the frames from the same subject tend to be closer in the feature space. On the other hand, as shown in (c) and (d), although the network is learned on the other dataset, fc7 features show great invariance to individual differences. More visualization of distributions over more AUs can be found in the supplementary material.

As a quantitative evaluation, we treated the frames from each subject as a distribution, and computed the distance between two subjects as Jensen-Shannon (JS) divergence [20]. Explicitly, we first compute a mean vector $\boldsymbol{\mu}_s$ for each subject $s$ in the feature space, and then squeeze $\boldsymbol{\mu}_s$ using a logistic function $\sigma(a) = \frac{1}{1+e^{-a/m}}$ ($m$ is median of $\boldsymbol{\mu}_s$ as the median heuristic) and unity normalization, so that each mean vector can be interpreted as a discrete probability distribution, *i.e.*, $\boldsymbol{\mu} \geq 0$, $\|\boldsymbol{\mu}\|_1 = 1$, $\forall s$. Given two subjects $p$ and $q$, we compute their JS divergence as:

$$D(\boldsymbol{\mu}_p, \boldsymbol{\mu}_q) = \frac{1}{2} D_{\text{KL}}(\boldsymbol{\mu}_p \| \mathbf{m}) + \frac{1}{2} D_{\text{KL}}(\boldsymbol{\mu}_q \| \mathbf{m}), \qquad (6)$$

where $\mathbf{m} = \frac{1}{2}(\boldsymbol{\mu}_p + \boldsymbol{\mu}_q)$ and $D_{\text{KL}}(\boldsymbol{\mu}_p, \mathbf{m})$ is the discrete KL divergence of $\boldsymbol{\mu}_p$ from $\mathbf{m}$. JS divergence is symmetric and smooth, and has been shown effective in measuring the dissimilarity between two distributions (*e.g.*, [36]). Higher value of $D(\boldsymbol{\mu}_p, \boldsymbol{\mu}_q)$ tells larger mismatch given the distributions for two subjects. Fig. 6 shows the divergences of each individual from two datasets. As can be seen, SIFT consistently reached a lower divergence than Gabor, providing an evidence that local descriptor (SIFT) is more robust to appearance changes compared to holistic ones (Gabor). This also serves as a possible explanation why SIFT consistently outperformed Gabor as found in [45]. Overall, fc7 yields much lower divergence compared to other popular engineered features.

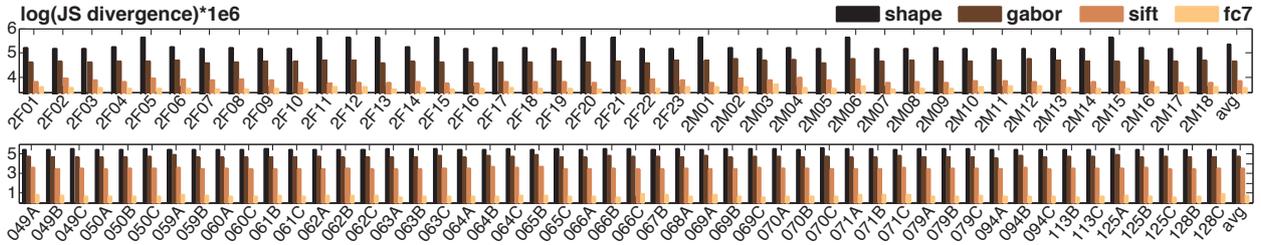

**Fig. 6.** Analysis of subject-invariance on two datasets: BP4D (**top row**) and GFT (**bottom row**). Four representative features, shape, Gabor, SIFT and fc7, were compared (details in text). For display purpose, a computed divergence $d$ is normalized by $\log(d) \times 1e6$.

### 4.4 Evaluation of detection performance

This section evaluates the performance of the proposed network on BP4D and GFT datasets. Below we summarize alternative methods, and then provide observations and discussion in hope to answer several fundamental questions.

**Alternative methods:** To evaluate the performance of the proposed network, we compared a baseline SIFT method, a standard multi-label CNN, and feature-based state-of-the-arts. The first alternative approach is a baseline SIFT approach, which has been shown to outperform other appearance descriptors (*i.e.*, Gabor/Daisy) [45]. Because SIFT is unsupervised, for fairness, we also evaluated a cross-dataset protocol to train AlexNet on the other dataset, termed as ANet$^{\mathsf{T}}$. ANet$^{\mathsf{T}}$ was then used to extract the fc7 features in comparison to SIFT descriptors. Linear SVMs were utilized as the base classifier, which also tells how separable different features are. That is, higher classification rate suggests an easier linear separation, which supports the idea that a good representation could reduce the burden of designing a sophisticated classifier. We evaluated ANet$^{\mathsf{T}}$ only on a 3-fold protocol, while we expect similar results could be obtained using 10-fold. Another alternative is our modified AlexNet (ANet), as mentioned in Sec. 3.1, with slightly different architecture and loss function (multi-label cross-entropy instead of multi-class softmax). ANet stood for a standard multi-label CNN, a representative of *feature learning* methods. On the other hand, CPM [41] and JPML [43] are feature-based state-of-the-art methods reported on the two datasets, while tackling the AU detection problem from different perspectives. CPM is one candidate method of *personalization*, which addresses the distributional shift in the feature space by progressively adapting a classifier to best separate a test subject. On the other hand, JPML models *AU correlations*, and meanwhile considers patch learning to select important facial patches for specific AUs. All experiments followed protocols as described in Sec. 4.2.

**Table 2.** F1 metrics on GFT dataset [4] using the 3-fold and the 10-fold protocol.

| AU | 3-fold protocol | | | | | cross | 10-fold protocol | | | | |
|---|---|---|---|---|---|---|---|---|---|---|---|
| | SIFT | CPM | JPML | ANet | Ours | ANet$^T$ | SIFT | CPM | JPML | ANet | Ours |
| 1 | 12.1 | 30.7 | 17.5 | **31.2** | **31.2** | 9.9 | 30.3 | 29.9 | 28.5 | 57.5 | **63.0** |
| 2 | 13.7 | 30.5 | 20.9 | 29.2 | **31.1** | 10.8 | 25.6 | 25.7 | 25.5 | 61.4 | **74.6** |
| 4 | 5.5 | – | 3.2 | **71.9** | 71.4 | 45.4 | – | – | – | **75.9** | 68.5 |
| 6 | 30.6 | 61.3 | **70.5** | 64.5 | 63.3 | 46.2 | 66.2 | 67.3 | 73.1 | 61.6 | **66.3** |
| 7 | 26.4 | 70.3 | 65.5 | 67.1 | **77.1** | 51.5 | 70.9 | 72.5 | 70.2 | **80.1** | 74.5 |
| 10 | 38.4 | 65.9 | **67.9** | 42.6 | 45.0 | 23.5 | 65.5 | 67.0 | 67.1 | 54.5 | **70.3** |
| 12 | 35.2 | 74.0 | 74.2 | 73.1 | **82.6** | 55.2 | 74.2 | 77.5 | 78.3 | **79.8** | 78.2 |
| 14 | 55.8 | **81.1** | 52.4 | 69.1 | 73.0 | 62.8 | 79.6 | 80.7 | 61.4 | **84.2** | 80.4 |
| 15 | 9.5 | 25.5 | 20.3 | 27.9 | **33.9** | 14.2 | 34.1 | 43.5 | 28.0 | 40.3 | **50.5** |
| 17 | 31.3 | 44.1 | 48.3 | 50.4 | **53.9** | 34.2 | 49.2 | 49.1 | 42.4 | 61.6 | **61.9** |
| 23 | 19.5 | 19.9 | 31.8 | 34.8 | **38.5** | 21.8 | 28.3 | 35.0 | 29.6 | 47.0 | **58.2** |
| 24 | 12.9 | 27.2 | 28.5 | **39.0** | 37.0 | 18.9 | 31.9 | 31.6 | 28.0 | **56.3** | 50.8 |
| Avg | 24.2 | 48.2 | 41.8 | 50.0 | **53.2** | 32.9 | 50.5 | 52.4 | 48.4 | 63.4 | **66.4** |

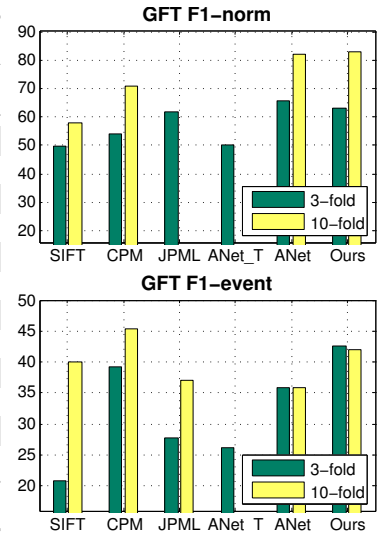

**Results and discussion:** Tables 2 and 3 show F1 metrics reported on 12 AUs; "Avg" for the mean score of all AUs. The bar plots show the averaged F1-norm and F1-event across all AUs. For detailed F1-frame and F1-event of individual AUs, please see supplementary materials. According to the results, we discuss our findings in hope to answer three fundamental questions:

1) *Could the learned representation generalize across subjects or datasets for AU detection?* On both datasets, compared to SIFT, ANet$^T$ trained with a cross-dataset protocol on average yielded higher scores with a few exceptions. In addition, for both 3-fold and 10-fold protocols where ANet was trained on exclusive subjects, ANet consistently outperformed SIFT over all AUs. These observations provide an encouraging evidence that the learned representation was transferable even when being tested across subjects and datasets, which also coincides with the findings in the image and video classification community [18, 31, 32, 39]. On the other hand, as can be seen, ANet trained within datasets leads to higher scores than ANet$^T$ trained across datasets. This is because of the dataset biases (*e.g.*, recording environment, subject background, etc.) that could cause distributional shifts in the feature space. In addition, due to the complexity of deep models, the performance gain of ANet trained on more data (10-fold) became larger than ANet trained on 3-fold, showing the generalizability of deep models increases with the growing number of training samples. Surprisingly, compared to SIFT trained on 10-fold, ANet trained on 3-fold showed comparable scores, even with ∼30% fewer data than what SIFT was used. All suggests that features less sensitive to the identity of subjects could improve AU detection performance.

2) *Could the learned temporal dependencies improve performance, and how?* The learned temporal dependencies was aggregated into the hybrid network denoted as "ours". On both 3-fold and 10-fold protocols, our hybrid network consistently outperformed ANet in all metrics. This improvement can be better told by comparing their F1-event scores. The proposed network used CNNs to extract spatial representations, stacked LSTMs to model temporal dependencies, and then performs a spatiotemporal fusion. From this view, predictions with fc7 features can be treated as a spacial case of ANet—a linear hyperplane with a portion of intermediate features. In general, adding temporal information helped predict AUs except for a few in GFT. A possible explanation is that in GFT, the head movement was more frequent and dramatic, and thus makes temporal modeling of AUs more difficult than moderate head movements in BP4D. In addition, adding temporal prediction into the fusion network attained an additional performance boost, leading to the highest F1 score on both datasets with either the 3-fold or the 10-fold protocols. This shows that the spatial and temporal cues are complementary, and thus is crucial to incorporate all of them into an AU detection system.

**Table 3.** F1 metrics on BP4D dataset [42] using the 3-fold and the 10-fold protocol.

| AU | 3-fold protocol | | | | | cross | 10-fold protocol | | | | | |
|---|---|---|---|---|---|---|---|---|---|---|---|---|
| | SIFT | CPM | JPML | ANet | Ours | ANet$^T$ | SIFT | CPM | JPML | ANet | Ours | |
| 1 | 21.1 | **43.4** | 32.6 | 40.3 | 31.4 | 32.7 | 46.0 | 46.6 | 33.9 | 54.7 | **70.3** | |
| 2 | 20.8 | **40.7** | 25.6 | 39.0 | 31.1 | 26.0 | 38.5 | 38.7 | 36.2 | 56.9 | **65.2** | |
| 4 | 29.7 | 43.3 | 37.4 | 41.7 | **71.4** | 29.0 | 48.5 | 46.5 | 42.2 | **83.4** | 83.1 | |
| 6 | 42.4 | 59.2 | 42.3 | 62.8 | **63.3** | 61.9 | 67.0 | 68.4 | 62.9 | 94.3 | **94.7** | |
| 7 | 42.5 | 61.3 | 50.5 | 54.2 | **77.1** | 59.4 | 72.2 | 73.8 | 69.9 | 93.0 | **93.2** | |
| 10 | 50.3 | 62.1 | 72.2 | **75.1** | 45.0 | 67.4 | 72.7 | 74.1 | 72.5 | 98.9 | **99.0** | |
| 12 | 52.5 | 68.5 | 74.1 | 78.1 | **82.6** | 76.2 | 83.6 | 84.6 | 72.0 | 94.4 | **96.5** | |
| 14 | 35.2 | 52.5 | 65.7 | 44.7 | **72.9** | 47.1 | 59.9 | 62.2 | 62.6 | 82.9 | 86.8 | |
| 15 | 21.5 | **36.7** | 38.1 | 32.9 | 34.0 | 21.7 | 41.1 | 44.3 | 38.2 | 55.4 | 63.3 | |
| 17 | 30.7 | **54.3** | 40.0 | 47.3 | 53.9 | 47.1 | 55.6 | 57.5 | 46.5 | 81.1 | 82.7 | |
| 23 | 20.3 | **39.5** | 30.4 | 27.3 | 38.6 | 21.6 | 40.8 | 41.7 | 38.3 | 63.7 | **73.5** | |
| 24 | 23.0 | 37.8 | 42.3 | **40.1** | 37.0 | 31.3 | 42.1 | 39.7 | 41.5 | 74.3 | **81.6** | |
| Avg | 32.5 | 50.0 | 45.9 | 48.6 | **53.2** | 43.4 | 55.7 | 56.5 | 51.4 | 77.8 | **82.5** | |

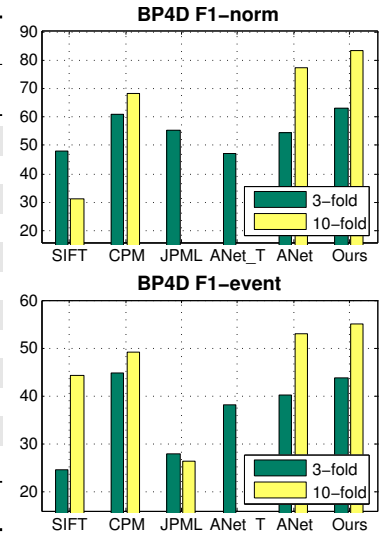

3) *Would jointly considering all issues in one framework improve AU detection?* This question aims to examine if the hybrid network would improve the performance of the methods that consider the aforementioned issues independently. To answer this question, we implemented CPM [41] as a personalization method that deals with representation issues, and JPML [43] as a multi-label learning method that deals with AU relations. Our modified ANet served as a feature learning method. All parameters settings were determined following the descriptions in the original papers. To draw a valid discussion, we fixed the exact subjects for all methods. Observing 3-fold on both datasets, the results are mixed. In GFT, ANet and JPML achieved 3 and 2 highest F1 scores; in BP4D, CPM and ANet reached 5 and 2 highest F1 scores. An explanation is because, although CNNs possess high degree of expressive power, the number training samples in 3-fold (33% left out for testing) were insufficient and might resulted in overfitting. In the 10-fold experiment, when training data was abundant, the improvements became clearer, as the parameters of the complex model can better fit our task. Overall, in most cases, the hybrid network outperformed alternative approaches by a significant margin, showing the benefits for jointly modeling all perspectives in one framework.

### 4.5    Visualization of learned AU models

To better understand and interpret how the proposed network produced good results, especially what input that flows through the network determines a specific AU, this section visualizes models of each AU learned by the CNN. Specifically, we implemented a gradient ascent technique [30,40] by synthesizing an input image that maximize acti-vations of AU units in the fc8 layer. More formally, we search such input image $\mathcal{I}^\star$ by solving the optimization problem:

$$\mathcal{I}^\star = \arg\max_{\mathcal{I}} A_\ell(\mathcal{I}) - \Omega(\mathcal{I}),  \tag{7}$$

where $A_\ell(\mathcal{I})$ is an activation function for the $\ell$-th unit of the fc8 layer given an image $\mathcal{I}$, and $\Omega(\cdot)$ is a regularization function that penalizes the image to enforce a natural image priors. In particular, we implemented $\Omega(\cdot)$ as a sequential operation of $L_2$ decay, clipping pixels with small norm, and Gaussian blur [40]. The optimization was done by iteratively updating a randomized and zero-centered image with the backprop gradient of $A_\ell(\mathcal{I})$. In other words, each pixel of $\mathcal{S}$ was renewed gradually to increase the activation of the $\ell$-th AU. This process continued until 10,000 iterations.

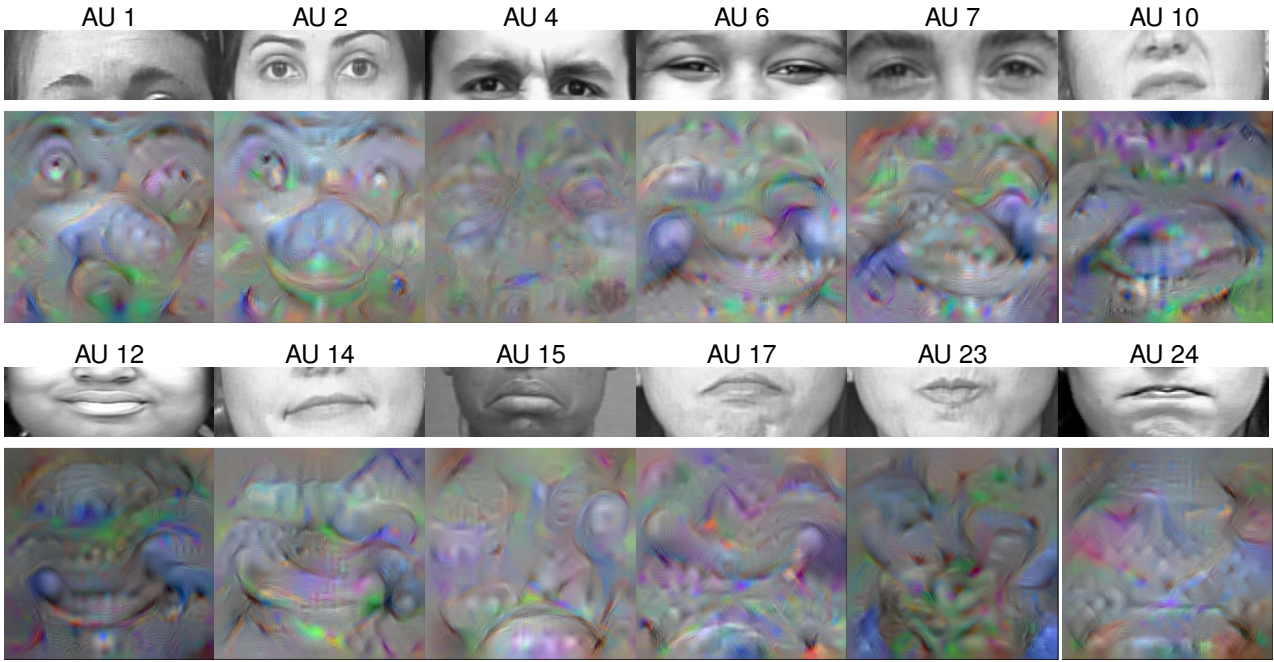

**Fig. 7.** Synthetically generated images to maximally activate individual AU neurons in the fc8 layer of CNN, trained on GFT [4], showing what each AU model "wants to see". The learned models show high agreement on attributes described in FACS [10]. (best view electronically)

Fig. 7 shows our visualizations of each AU model learned by the CNN architecture described in Sec. 3.1. As can be seen, most models match the attributes described in FACS [10]. For instance, model AU12 (lip corner puller) exhibits a strong "⌣" shape to the mouth, overlapped with some vertical "stripes", implying that the appearance of teeth is commonly seen in AU12. Model AU14 (dimpler) shows the dimple-like wrinkle beyond the lip corner, which, compared to AU12, gives the corners of the lips a downward cast. Model AU15 (lip corner depressor) shows a clear "⌢" shape to the mouth, producing an angled-down shape at the corner. For upper face AUs, model AU6 (cheek raiser) captures deep texture of raised-up cheeks, narrowed eyes, as well as a slight "⌣" shape to the mouth, suggesting its frequent co-occurrence with AU12 in spontaneous smiles. Models AU1 and AU2 (inner/outer brow raiser) both capture the arched shapes to the eyebrows, horizontal wrinkles above eyebrows, as well as the widen eye cover that are stretched upwards. Model AU4 (brow lowerer) captures the vertical wrinkles between the eyebrows and narrowed eye cover that folds downwards.

Our visualizations suggest that the CNN was able to identify these important spatial cues to discriminate AUs, even though we did not ask the network to specifically learn these AU attributes. In addition, the global structure of a face was actually preserved throughout the network, despite that convolutional layers were designed for local abstraction (*e.g.*, corners and edges as shown in Fig. 3). The widespread agreements between the synthetic images and FACS [10] confirm that the learned representation is able to describe, and thus reveal these attributes across multiple AUs. This was not shown possible in standard hand-crafted features in AU detection (*e.g.*, shape [15, 24], SIFT [41, 43], LBP [17, 35], or Gabor [35]). To the best of our knowledge, this is the first time to visualize how machines see facial AUs.

## 5   Conclusion and future work

In this paper, we have presented a hybrid network architecture that jointly models three issues arising in AU detection: *Spatial representation*, *temporal modeling*, and *AU correlation*. To the best of our knowledge, this is the first study that shows a possibility for exploring the three seemingly unrelated aspects within one framework. The hybrid network is motivated by existing progress on deep models, and takes advantage of spatial CNNs, temporal LSTMs, and their fusions to achieve multi-label AU detection. In particular, compared to popular hand-crafted features in AU detection, we empirically showed that a spatial representation can be learned, reduces sensitivity to the identity of subjects, and further improves performance even with a linear classifier. Experiments on two of the largest spontaneous AU datasets demonstrate that the proposed network outperformed a standard CNN and feature-based state-of-the-art methods. In addition, our visualization of learned AU models showed, for the first time, how machines interpret facial AUs. Such visualization provides strong evidence for learning discriminative AU features and AU correlations. Future work include deeper investigation/analysis of this hybrid network, and incorporation of bi-directional LSTMs. More comparisons can be also considered, *e.g.*, keeping LSTM but replacing fc7 with SIFT, or keeping fc7 but replacing LSTM with HMM/CRF.